\documentclass[11pt,a4paper]{article}
\usepackage[utf8]{inputenc}
\usepackage[OT1]{fontenc}
\usepackage[english]{babel}
\usepackage[hyperref]{acl2019}
\usepackage{times}
\usepackage{latexsym}
\usepackage{booktabs}

\PassOptionsToPackage{hyphens}{url}\usepackage{hyperref}

\aclfinalcopy 

\title{Lingua Custodia at WMT'19: \\Attempts to Control Terminology}

\author{Franck Burlot \\ %
  Lingua Custodia \\
  Montigny-le-Bretonneux, France \\
  {\tt\small franck.burlot@linguacustodia.com} }

\date{}

\begin{document}
\maketitle
\begin{abstract}
  This paper describes Lingua Custodia's submission to the WMT'19
  news shared task for German-to-French on the topic of the EU
  elections. We report experiments on the adaptation of
  the terminology of a machine translation system to a specific topic,
  aimed at providing more accurate translations of specific entities like
  political parties and person names, given that
  the shared task provided no in-domain training parallel data dealing
  with the restricted topic.
  Our primary submission to the shared task
  uses backtranslation 
  generated with a type of decoding allowing
  the insertion of constraints in the output
  in order to guarantee the correct translation of specific terms
  that are not necessarily observed in the data.
\end{abstract}

\section{Introduction}

A sub-task of the WMT'19 News Translation
shared task has been jointly organized by the University of Le Mans
and Lingua Custodia: the translation of news articles dealing
with the topic of the 2019 European Parliament elections for
the French-German language pair. This brings back French,
a language absent from the News Translation task since 2015,
and pairs it with German, a morphologically richer language
than English. Finally, the EU election topic brings new
challenges to the task.

Such a restriction of the domain to a single topic makes
the task very different from the translation of any news
data. We propose to roughly define a domain according to two
majors dimensions:

\begin{itemize}
\item \textbf{Syntactic structure}. The European election topic
  probably has no or few syntactic and stylistic differences with
  the general news domain, since we are in both cases dealing with
  news articles with the same characteristics. On the other hand,
  sentences in newspapers are generally longer than in casual discourse.
\item \textbf{Terminology}. A specific topic implies a specific
  terminology. For instance, the system should not attempt a literal
  translation of the German politician's name {\it Wagenknecht}. It
  should also be aware of the specific translations of political party names
  in the press of the target language: the French party {\it France Insoumise}
  should not be translated into German. Furthermore, the French
  movement {\it gilets jaunes} (yellow vests) is refered to in the German press as {\it Gelbwesten},
  and a literal translation, such as {\it gelbe Westen}, is inaccurate.
\end{itemize}

There exist efficient methods for domain adaptation in neural MT
\cite{Luong-Manning:iwslt15,chu-wang-2018-survey}. The experiments
introduced in this paper attempt to explore techniques that help
to specifically adapt the terminology of a system to a restricted topic.
However, a serious difficulty stands in the way: among the parallel data
provided for the task, only 1,701 sentence pairs deal with the EU elections
(development set).
Recent monolingual data in German and French is available and contains
several sentences using the required terminology, but we then lack
the correct translations of the terms of interest.

This paper describes Lingua Custodia's attempts to specifically control
the terminology generated by a Machine Translation (MT) system, using
only the data provided at the Conference.
The resulting German-to-French system was submitted at WMT'19.

In the first section, we provide an overview of our baselines and
point out several terminology issues. We then
describe our experiments with constrained decoding to control terminology.
The last section introduces an attempt to relax the hard constraints
applied to the decoder.

\section{Baseline \label{sec:baseline}}

The training parallel data provided for the task consisted of
nearly 10M sentences, including {\it Europarl} \cite{Koehn05europarl},
{\it Common-crawl}, {\it News-commentary} and {\it Bicleaner07}. The former was
the biggest (over 7M sentences) and also the noisiest corpus,
containing bad characters, short phrases with only numbers,
lists of products, sentences in the wrong language,
obviously machine translated sentences, etc.

\subsection{Data selection \label{subsec:data}}

We have performed a filtering of the {\it Bicleaner07} corpus
in order to reduce the impact of noisy samples on the MT system,
using {\it LC\_Pruner}, a in-house system that was submitted at the
First Automatic Translation Memory Cleaning Shared Task
\cite{BarbuEBNTOF16}.
The system extracts several monolingual and bilingual features
that are fed to a random forest classifier aimed at predicting
if a sentence pair is a good translation and whether each sentence
is well formed. It is based on the following features:

\begin{itemize}
  \item Total sentence pair length
  \item Source/target length ratio
  \item Average token length
  \item Uppercase token count comparison
  \item Source/target punctuation comparison
  \item Source/target number comparison
  \item Language identification using {\it langid.py} \cite{lui-baldwin-2012-langid}
  \item Cognates
  \item Source and target language model scores
  \item Hunalign scores \cite{varga07hunalign}
  \item Zipporah adequacy scores \cite{xu-koehn-2017-zipporah},
    using a probabilistic bilingual dictionary computed on Europarl.
\end{itemize}

Random forest parameters are optimized using expert feedback
on a set of parallel sentences automatically selected by
the model across several iterations. We have run 3 iterations,
assessing the quality of 20 sentence pairs each time. The result
is a binary classification of each sentence pair based on
a score between 0 and 1. We have experimented
with two selection criteria, keeping sentence pairs scoring above 0.5
and above 0.8, which led to respectively nearly 4M and 2M
finally accepted sentences. The results are introduced in Section~\ref{blRes}.

\subsection{System setup \label{setup}}

German and French pre-processing was performed using in-house normalization
and tokenization tools. Truecasing models were learnt, using Moses scripts \cite{Koehn07moses}, on the monolingual news
data provided at the Conference, on all 2017-2018 data for French and 10M sentences
from 2018 for German. A shared French-German BPE
vocabulary \cite{Sennrich16BPE} was built with 30k merge operations
on all the parallel data available for the task, except {\it Bicleaner07}.

We have trained baseline systems for French-German in both directions.
Transformer {\it base} \cite{vaswani17attention} models were trained using
the Sockeye toolkit \cite{hieber17sockeye} on two Nvidia 1080Ti GPU cards.
Most of the standard hyper-parameters have been used.  The model dimension included 512 units.
The initial learning rate was set to $0.0003$ with a warmup on for $30k$ updates.
Due to the small quantity of training data available, we decided to
slightly increase dropout between layers ($0.2$) and label smoothing ($0.2$).
Validations were performed every $20k$ updates and patience was set to $15$.
Since this setup contained no training data relevant to the EU election
topic, we decided to hold out the provided development set for another purpose,
and used a general news domain test set: {\it Newstest-2012}.
We finally wished to sample more sentence pairs from news-related corpora
during training. Since no such method is implemented in
the Sockeye toolkit for minibatch generation, we simply trained
the baselines on a single copy of {\it Bicleaner07} and {\it Common-crawl},
and took two copies of {\it Europarl} and 6 of {\it News-commentary}.

\subsection{Results and terminology issues \label{blRes}}

The systems were tested on the official development set,
{\it Euelections-dev-2019}, as well as {\it Newstest-2013}
and the official test set {\it Newstest-2019}. BLEU scores were computed
with {\it SacreBLEU} \cite{post-2018-call} and are shown in
Table~\ref{tab:blRes}.

\begin{table*}
\begin{center} \small
\begin{tabular}{l|ccc}
\hline
\multicolumn{4}{c}{French-to-German} \\
\hline
& Euelections-dev-2019 & Newstest-2013 & Newstest-2019 \\
\textbf{Baseline} & 25.98 & 23.48 & 26.94 \\
\hline
\multicolumn{4}{c}{German-to-French} \\
\hline
& Euelections-dev-2019 & Newstest-2013 & Newstest-2019 \\
\textbf{LC\_Pruner $2M$} & 31.07 & 27.49 & 33.04 \\
\textbf{LC\_Pruner $4M$} & 30.96 & 27.29 & 33.16 \\
\hline
\end{tabular}
\end{center}
  \caption{BLEU scores for French-German baselines}
  \label{tab:blRes}
\end{table*}

Experiments with different data filtering criteria for
the {\it Bicleaner07} corpus were introduced in subsection~\ref{subsec:data}.
We observe that keeping a bigger set of data does not lead
to any clear improvements, at least in terms of BLEU.
Thus we have {\it kept LC\_Pruner $2M$} as the main
baseline for further training in Section~\ref{subsec:synth}.

The translation from English into German of {\it Euelections-dev-2019}
by our baseline shows consistent terminology
issues. The systems has difficulties translating the name of
the movement {\it gilets jaunes} (yellow vests). Out of the
19 occurrences of the expression in the French source, only
4 are correctly translated as the compound {\it Gelbwesten}. We noted
several translations as {\it gelbe Westen}, the translation
of the adjective {\it jaunes} only, as well as full omissions.
We also noted that the French party {\it France Insoumise}
was translated litterally as {\it unbeugsame Frankreich},
instead of simply being copied, the name of the politician
{\it Nicolas Dupont-Aignan} was translated as {\it Nicolas Du\textbf{m}ont-Aignan}, etc.
Our best baseline translates the German side of this test
into French with the same kind of difficulties: {\it Gelbwesten}
is sometimes translated as {\it la veste jaune}, etc.

\section{Terminology control}

We argue that a system specialized in a specific topic
should be able to provide the right translations for
terms that are relevant to this topic. The baselines
we have just introduced fail to translate important
terminology. We now seek to adapt these baselines to
the EU election terminology.

\subsection{Constrained decoding \label{constrDec}}

One way to integrate such knowledge of a specific terminology
into the MT system is by using {\it constrained decoding} \cite{hokamp-liu-2017-lexically}.
The {\it Grid Beam Search} algorithm guarantees
the presence of one or several given phrases in the MT output.
This method does not require any change in the
model or its parameters, thus the algorithm does not model any
sort of token-level source-to-target relation, but simply forces the beam search
to go through the target constraint. The challenge for the decoder
is then to correctly insert the constrained phrase in the rest
of the sentence.

\citet{post-vilar-2018-fast} proposed a variant of this
algorithm with a significant lower computational complexity.
We used their implementation available in the
Sockeye toolkit.

\subsection{Lexicon extraction}

We have extracted bilingual lexicons from two sources: the official
development set provided for the task ({\it Euelections-dev-2019}),
and the monolingual French and German data made available at WMT.

\subsubsection{Parallel EU election data}

We have decided to use the official development set ({\it Euelections-dev-2019}) as the main
source of terminology, for the simple reason that it is the
only parallel data available containing the specific terminology
of the EU elections with reliable human translations.

Alignments were learnt using Fastalign \cite{dyer-etal-2013-simple}
on a concatenation of {\it News-commentary} and {\it Euelections-dev-2019}, and we used
them to extract a phrase table from the former with the Moses toolkit.
We removed a phrase pair whenever the probability of the German side,
given the French side, was below $0.5$. This ensured that we never keep
more than one translation for a French phrase\footnote{Since there can be several
French translations for one German phrase, the current terminology
can only be used for translation into German.}.

The resulting phrases were furthermore filtered according to their
domain. We computed Moore-Lewis \cite{moore-lewis} scores of the source French phrases.
The out-of-domain language model was computed on the French side
of the parallel data (section~\ref{sec:baseline}), and the in-domain
model on the French monolingual news data 2018 available at WMT. Although
this corpus does not contain exclusively articles about the EU
elections, we believe its terminology distribution may be closer
to what is observed in {\it Euelections-dev-2019}, because
the corpus relates more recent news. We kept the best 2000 phrase
pairs according to their Moore-Lewis score.

Finally, we kept the phrase pairs for which the German side appeared
at least once in the German monolingual news 2018 corpus, in order
to filter out obviously bad expressions that remained. We ended
up with 773 phrase pairs, among which could be found the correct
translation of {\it gilets jaunes} (yellow vests).

\subsubsection{Monolingual news data}

As an attempt to address the issue of person name mistranslations,
we extracted named entities from the French monolingual news 2018 corpus.
First, we tagged the corpus with an in-house French named entity recognizer. 
We then computed the tagged named entity occurrence counts over the same corpus
and removed the ones occurring less than 9 times. The translations of
the extracted expressions into German are unknown,
so we looked for the named entities that are not translated, but
copied into German. We therefore kept the entries that had an
occurrence count higher than 9 in the German news monolingual 2018
corpus. As a result, the name {\it Poutine} in French would be
removed because it translates into a different word in German
({\it Putin}), whereas {\it Dupont-Aignan} would be kept, as it
stays the same in both languages. This procedure produced
nearly 20k phrase pairs.

Prior to inference, constraints extracted from the development set
are applied every time a source-side constraint is found
in the source sentence to be translated. Named Entity constraints
extracted from monolingual data are applied in a different way.
The same named entity classifier as above is used to tag the source
sentence and a constraint is applied when: 1. the source constraint
matches a part of the sentence ; 2. the matched sentence part has
been tagged as a named entity.

We are well aware that bilingual terminology extraction is a complex
task and that more sophisticated models need to be investigated.
We chose to employ these simple heuristics only because we lacked time.
We did run experiments with tools, allowing us to extract bilingual lexicons
from monolingual data, namely {\it Muse} \cite{conneau17word}
and {\it BiLex} \cite{zhang17bilex}. However, we found
them not suited for our requirements, because 1. the global quality
of the lexicons was too low to be inserted in a MT decoder as hard constraints,
and 2. only single-word phrases were extracted
and we wished to extract multi-word expressions as well.
Future work should include methods for phrase pair extraction
from monolingual data \cite{marie18phrase,artetxe19unsup}.

\subsection{Constrained French-to-German baseline \label{fr-deRes}}

The scores of the French-to-German baseline with and without
constraints are shown in Table~\ref{tab:constRes}.
We used a beam size of 20
for constrained decoding, as recommended in the Sockeye documentation
\footnote{\url{https://awslabs.github.io/sockeye/inference.html}},
and a default beam size of 5 for the unconstrained decoding. 
The final models are averages of the 4 best checkpoints in terms of BLEU
on the validation set. Applying
constraints to {\it Euelections-dev-2019} adds 2 BLEU points to the baseline,
but this should not be considered as an improvement, since
parts of the reference translations were inserted as constraints.
We observe that constrained decoding has nearly no impact
on the BLEU score for {\it Newstest-2013}, and that it even slightly
degrades the score for {\it Newstest-2019}.

\begin{table*}
\begin{center} \small
\begin{tabular}{l|ccc}
\hline
& Euelections-dev-2019 & Newstest-2013 & Newstest-2019 \\
\textbf{Baseline} & 25.98 & 23.48 & 26.94 \\
+ Constraints     & 27.87 & 23.42 & 26.66 \\
\hline
\end{tabular}
\end{center}
  \caption{BLEU scores for French-to-German with constrained decoding}
  \label{tab:constRes}
\end{table*}

\begin{table*}
\begin{center} \scriptsize
\begin{tabular}{l|l}
\toprule
Source & Même si les populistes de gauche ont bien moins de succès en Europe que les acteurs d’extrême-droite, ils peuvent encore s’imposer, \\
& comme le montre l’ascension de partis classiques d’opposition tels que \textbf{Podemos} en Espagne et \textbf{La France Insoumise} en France. \\
\midrule
Constraints & Podemos, France Insoumise \\
\midrule
English & Even if left-wing populists have far less success in Europe than right-wing actors, they can still prevail, as evidenced by the rise of \\
& classic opposition parties such as \textbf{Podemos} in Spain and \textbf{France Insoumise} in France. \\
\midrule
Baseline & Obwohl die Linkspopulisten in Europa deutlich weniger erfolgreich sind als die Rechtsextremen, können sie sich immer noch durchsetzen, \\
& wie der Aufstieg klassischer Oppositionsparteien wie \textbf{Podemos} in Spanien und Frankreichs \textbf{Ununterwürfiges Frankreich} zeigt. \\
\midrule
+ Constraints & \textbf{Podemos} in Spanien und \textbf{France Insoumise} in Frankreich haben zwar deutlich weniger Erfolg als rechtsextreme Populisten,\\
& aber sie können sich noch immer durchsetzen. \\
\midrule
Reference & Auch wenn die Linkspopulisten in Europa weitaus weniger erfolgreich sind als die Rechts-außen-Player, können sie sich durchaus Geltung\\
& verschaffen, wie der Aufstieg klassischer Herausforderer-Parteien wie \textbf{Podemos} in Spanien und \textbf{La France Insoumise} in Frankreich zeigt. \\
\bottomrule
\end{tabular}
\end{center}
  \caption{Example of French-to-German translation with and without constrained decoding ({\it Newstest-2019})}
  \label{tab:exConstr}
\end{table*}

The low impact of the constraints on {\it Newstest-2013} may be explained
by the fact that this set is irrelevant with regard to the EU election topic,
leading to the insertion of few constraints: 465 constraints were inserted
in 3000 sentences. As a comparison, 751 constraints were inserted in the 1701
sentences of {\it Newstest-2019}. Looking more closely at the outputs of
the different systems, we observed several cases where : 1. the constraint was
erroneously inserted in the sentence; 2. the insertion of a constraint
seemed to disturb the decoder, which resulted in broken sentences. 
Table~\ref{tab:exConstr} illustrates a case where the constraint helped to
correct a mistranslation, but both issues occurred. The French party {\it France Insoumise}
was translated litterally by the baseline into {\it Ununterwürfiges Frankreich},
and one of our constraints successfully forced the right translation of this expression.
First, the subject of the first clause
({\it les populistes de gauche}) has been replaced by the constraints, which
should have been inserted in the end of the sentence, like in the baseline.
Second, the constrained output ignores the whole section about the raise of
classical populist parties.

Although several constraints may potentially help the adaptation
of a MT system to the specific terminology of the EU elections,
it may be possible that the positive impact it could have on BLEU is
mitigated by the broken translations the constraints tend to produce.

\section{Relaxed use of constraints}

We assume that the strict insertion of terminology through
constrained decoding sometimes breaks output sentences, partly
because the decoder would have never generated such an
expression by itself. More specifically, the decoder assigns
a low probability to the constrained phrase, which leads to
a harmful disruption during the beam search.

Using parallel data containing the required terminology to fine-tune a system
is an obvious good way to adapt a system, and it has the advantage
to leave the decoder unchanged. Although we have no such data
available for training, we do have monolingual French data
that contains at least a big part of the EU election terminology
we wish to acquire: the monolingual news 2018 corpus released
within the shared task. We could use our French-to-German baseline
to backtranslate these sentences \cite{Sennrich16improving},
but this would have the effect of introducing mistranslations
in the source, which would break the strict source-target
mapping we need to learn. For instance, if the French phrase
{\it gilets jaunes} is backtranslated as {\it gelbe Westen},
the final German-to-French system would learn to translate {\it gelbe Westen} into French,
but could very well still produce erroneous translations of
the correct source expression {\it Gelbwesten}.

To address this issue, we propose to apply the strict
constraints (section~\ref{constrDec}) to the French-to-German
baseline used for backtranslation. Although we condemn ourselves
to certain broken translated outputs, we have the guarantee
that the extracted constraints will be learnt by
the system. Another advantage of this strategy is that the
constraints are inserted in different contexts, which should help the decoder
learn to insert constrained terms in the output sentences more correctly.

\subsection{Synthetic parallel datasets \label{subsec:synth}}

The French news monolingual corpus 2018 comes under the general news
domain. We attempted to extract the sentences dealing with the EU election
topic using Moore-Lewis data selection strategy \cite{moore-lewis}. We chose
the French side of {\it Euelections-dev-2019} as our in-domain corpus,
with the hope that it will favor sentences containing the constraints
we have extracted from it, in order to maximize the presence of
constraint pairs in the backtranslated data. We finally selected
the best 2M sentences in terms of Moore-Lewis score.

We provide both constrained and unconstrained translations
for the resulting French sentences, using the same beam sizes
as in Section~\ref{fr-deRes}. The constrained setup inserted
673,670 phrases in 2M German sentences.

\subsection{Results}

\begin{table*}
\begin{center} \small
\begin{tabular}{l|ccc}
\hline
& Euelections-dev-2019 & Newstest-2013 & Newstest-2019 \\
\textbf{Baseline}                      & 31.07 & 27.49 & 33.04 \\
\textbf{Unconstrained} & 34.06 & 28.07 & 35.64 \\
\textbf{Constrained}   & 34.04 & 27.99 & 35.45 \\
\textbf{Ensemble}                 & 34.31 & 28.10 & 35.62 \\
\hline
\end{tabular}
\end{center}
  \caption{BLEU scores for German-to-French systems fine-tuned on backtranslated data}
  \label{tab:finalRes}
\end{table*}

We used the German-to-French baseline trained on 2M sentences
from {\it Bicleaner07} (section~\ref{blRes}) as a starting point
for fine-tuning using the constrained and unconstrained versions
of the backtranslation. The backtranslated data was mixed with
{\it Europarl} and {\it News-commentary} corpora.
We first tried to use {\it Newstest-2012} for validation, but only a slight
improvement was observed throughout the training in terms of BLEU.
In order to avoid stopping the training too early, we finally
decided to run validation on {\it Euelections-dev-2019}. This
most certainly led to overestimated BLEU scores, since the
backtranslation data has been selected according to its proximity
to this development set (section~\ref{subsec:synth}). However,
it allowed the stopping criterion to fire later during training.

The final models we introduce are averages of the 4 best checkpoints
in terms of BLEU on {\it Euelections-dev-2019}. We also provide
results for an ensemble of 8 checkpoints (4 best constrained and
4 best unconstrained).
We kept the same hyper-parameters as
described in Section~\ref{setup}, except we lowered the learning
rate from 0.0003 to 0.0001, used no warmup, and ran more frequent
validations (every 10k updates).

The result of these fine-tuning procedures are shown in Table~\ref{tab:finalRes}.
Both backtranslation setups provide the best improvements we observed
on {\it Newstest-2019} (~+2.5). However, we see no significant
difference between the constrained and unconstrained setups.
This could be expected, since our experiment was focused
on a small set of terms we wished the systems to generate, which
can only lead to local improvements with low impact on the BLEU score.
The ensemble of 8 models combining both setups is our primary submission to the shared task.

We have run a small analysis of the outputs given by both setups
for {\it Newstest-2019}. We observed that the constrained system correctly copied the
German name {\it Alexander Gauland}\footnote{Constraint: {\it Alexander Gauland} $\rightarrow$ {\it Alexander Gauland}}, whereas the unconstrained system
erroneously translated the first name into {\it Alexandre}. The constrained
system also translated {\it europäischen Vermögenssteuer} (European wealth tax)
into the acronym {\it ISF européen}\footnote{Constraint: {\it Vermögenssteuer} $\rightarrow$ {\it ISF}}, which seems more usual in the press
about the EU elections, compared to the litteral translation of the
unconstrained system as {\it impôt européen sur la fortune}.
Several phrases that were in our extracted constraints were correctly
translated by the unconstrained system as well. Unconstrained backtranslation
\cite{Sennrich16improving} thus seems to be sufficient to adapt the
terminology of a system to a specific system, at least in our setup
with few low-quality automatically extracted lexical constraints.
However, both systems produce consistent errors on terms that we
failed to capture in constraints, which leads us to think that
higher quality constraints should have a bigger positive impact
on terminology adaptation.

\section{Conclusions}

We have described Lingua Custodia's submission to WMT'19
News Translation shared task. We attempted to adapt the
terminology of a MT system to the EU election topic without
relevant parallel training data. Forcing the decoder to generate
specific terms can help, although it disturbs the decoder, which may lead to broken output sentences.
Using hard constraint insertion to generate backtranslated target
monolingual data showed no improvement in terms of BLEU scores,
but we have observed local improvements in the generated terminology.
The system that has been submitted to the shared task is an
ensemble of both constrained and unconstrained models.

Lexically constrained decoding is highly dependent on the quality
of the bilingual constraints available. In future work, we plan to
search for other techniques for automatic lexical constraint extraction
in order to improve recall and reach a better terminology coverage.
We also plan to investigate new techniques to relax the hard
constraints applied to the decoder, in order to impose less
disturbance to the beam search and avoid broken output sentences.


\bibliography{biblio}
\bibliographystyle{acl_natbib}

\end{document}